\documentclass[conference]{IEEEtran}
\IEEEoverridecommandlockouts
\usepackage{cite}
\usepackage{amsmath,amssymb,amsfonts}
\usepackage{algorithmic}
\usepackage{graphicx}
\usepackage{textcomp}
\usepackage{xcolor}
\usepackage{bm}

\def\BibTeX{{\rm B\kern-.05em{\sc i\kern-.025em b}\kern-.08em
    T\kern-.1667em\lower.7ex\hbox{E}\kern-.125emX}}

\begin{document}

\title{Deep activity propagation via weight initialization in spiking neural networks}


\author{\IEEEauthorblockN{1\textsuperscript{st} Aurora Micheli}
\IEEEauthorblockA{
\textit{TU Delft}\\
Delft, the Netherlands \\
a.micheli@tudelft.nl}
\and
\IEEEauthorblockN{2\textsuperscript{nd} Olaf Booij}
\IEEEauthorblockA{
\textit{TU Delft}\\
Delft, the Netherlands \\
o.booij@tudelft.nl}
\and
\IEEEauthorblockN{3\textsuperscript{rd} Jan van Gemert}
\IEEEauthorblockA{
\textit{TU Delft}\\
Delft, the Netherlands \\
j.c.vangemert@tudelft.nl}
\and
\IEEEauthorblockN{4\textsuperscript{th} Nergis T\"omen}
\IEEEauthorblockA{
\textit{TU Delft}\\
Delft, the Netherlands \\
n.tomen@tudelft.nl}
}

\maketitle

\begin{abstract}
Spiking Neural Networks (SNNs) and neuromorphic computing offer bio-inspired advantages such as sparsity and ultra-low power consumption, providing a promising alternative to conventional artificial neural networks (ANNs). However, training deep SNNs from scratch remains a challenge, as SNNs process and transmit information by quantizing the real-valued membrane potentials into binary spikes. This can lead to information loss and vanishing spikes in deeper layers, impeding effective training. While weight initialization is known to be critical for training deep neural networks, what constitutes an effective initial state for a deep SNN is not well-understood. Existing weight initialization methods designed for ANNs are often applied to SNNs without accounting for their distinct computational properties. In this work we derive an optimal weight initialization method specifically tailored for SNNs, taking into account the quantization operation. We show theoretically that, unlike standard approaches, our method enables the propagation of activity in deep SNNs without loss of spikes. We demonstrate this behavior in numerical simulations of SNNs with up to 100 layers across multiple time steps. We present an in-depth analysis of the numerical conditions, regarding layer width and neuron hyperparameters, which are necessary to accurately apply our theoretical findings. Furthermore, we present extensive comparisons of our method with previously established baseline initializations for deep ANNs and SNNs. Our experiments on four different datasets demonstrate higher accuracy and faster convergence when using our proposed weight initialization scheme. Finally, we show that our method is robust against variations in several network and neuron hyperparameters.
\end{abstract}


\section{Introduction}
Spiking Neural Networks (SNNs) are a class of artificial neural networks inspired by the dynamics of biological brains, where information is encoded and transmitted through discrete action potentials, or spikes~\cite{Abbott1999Lapicques1907,Hunsberger2015SpikingNeurons,Maass1997NetworksModels}. This unique mode of communication enables SNNs to perform fast computations with remarkably low power consumption~\cite{Indiveri2011NeuromorphicCircuits,Maass2004OnNeurons}, especially when combined with specialized neuromorphic hardware~\cite{Davies2018Loihi:Learning,Akopyan2015TrueNorth:Chip,Yamazaki2022SpikingReview}. However, the task performance of SNNs is still not comparable with that of conventional artificial neural networks (ANNs). This discrepancy can be ascribed to the additional challenges associated with their training. Relying on the differentiability of the loss function with respect to the network parameters, ANNs are typically trained using a gradient descent algorithm. However, the discrete nature of spikes prevents the direct use of backpropagation in SNNs. Different methods such as ANN-SNN conversion~\cite{Deng2021OptimalNetworks,Ding2021OptimalNetworks,Ho2020TCL:Layers} and backpropagation with surrogate functions~\cite{Neftci2019SurrogateNetworks,Zenke2021TheNetworks} are proposed to circumvent this problem. Nevertheless, the proposed solutions haven't been sufficient to fully bridge the accuracy gap between SNNs and ANNs, without compromising the efficiency advantages of SNNs. Therefore, novel perspectives on other aspects of SNN optimization are necessary.

Similarly to ANNs, in SNNs a suboptimal weight initialization can lead to vanishing or exploding gradients and hamper the training process~\cite{Hochreiter1997LongMemory,Rossbroich2022Fluctuation-drivenTraining}. This issue becomes even more evident in deep networks~\cite{Lee2016TrainingBackpropagation}. The question of how to properly initialize the weights in a network has been widely explored in the ANN literature, and different initialization strategies have been proposed tailored to specific activation functions and weight distributions~\cite{GlorotUnderstandingNetworks,He2015DelvingClassification,Mishkin2015AllInit}.
On the other hand, SNNs are often initialized following standard schemes designed for ANNs, irrespective of their distinct properties. Unlike ANNs, SNNs feature temporal dynamics, resetting mechanisms, information quantization and their activation function differs from those examined in the ANN literature. Hence, ANN initalization schemes are inadequate for SNNs and often lead to undesired effects such as vanishing or exploding spikes in deeper layers. In this paper:
\begin{itemize}[]
  \item We analytically derive a weight initialization method which takes into account the specific activation function of a Spiking Neural Network, following the variance flow approach proposed in~\cite{He2015DelvingClassification} for standard ANNs.
  \item We show that, unlike the standard method for the Rectified Linear Unit (ReLU) activation function, our initialization enables spiking activity to propagate from the input to the output layers of deep networks without being dissipated or amplified. 
  \item We empirically validate our theoretical findings on simulations of deep SNNs up to 100 layers and 20 time steps and present an in-depth analysis on the impact of neuron and network hyperparameters on the results.
  \item We compare our initialization method against 7 baseline methods on four distinct datasets and two network architectures, and we find that our proposed method leads to better accuracy, faster convergence and lower latency.
\end{itemize}

\section{Related work}
A proper initialization method should avoid reducing or amplifying
the magnitudes of input signals across layers. In the context of ANNs, Glorot \& Bengio~\cite{GlorotUnderstandingNetworks} address saturated units with the logistic sigmoid activation function and propose a weight initialization scheme aimed at maintaining constant activations and gradient variance across layers. He et. al~\cite{He2015DelvingClassification} extend this analysis to include the ReLU non-linearity and introduce what is conventionally called the Kaiming initialization, now widely adopted in deep ANNs.
In comparison, there are fewer studies on initialisation schemes in SNNs. Some research regards ANN-SNN conversion as an initialization method for SNNs~\cite{Rathi2020EnablingBackpropagation}, which confines SNN design to traditional ANN architectures and limits the exploration of alternative network structures. In addition, conversion methods are limited to using a single, rate-based encoding strategy and do not allow for networks with arbitrary encoding schemes to be trained from scratch, limiting SNNs flexibility and adaptability.

The problem of how information propagates in SNNs has been indirectly addressed. In~\cite{Rathi2020DIET-SNN:Networks} and~\cite{Zimmer2019TechnicalPyTorch} the appropriate membrane leak factor and firing threshold are learned during training. Similarly, in~\cite{Yin2020EffectiveNetworks,Fang2021IncorporatingNetworks} some learnable parameters are incorporated in the SNN to optimize the neuron firing rate. However, this approach inevitably results in increased computational complexity, which can hinder the scalability and efficiency of the model. A different way to control the neurons' output is to regularize the spike response. For instance,~\cite{Kim2021OptimizingSensing}~introduces a global unsupervised firing rate normalization. Similarly, \cite{Zheng2021GoingNetworks} and~\cite{Kim2020RevisitingScratch} adapt the concept of batch normalization~\cite{Ioffe2015BatchShift} to SNNs and propose a batch normalization method along channel and time axes, respectively. This approach aims to stabilize the firing rates during training, analogous to how BN stabilizes activations in ANNs, ensuring smoother gradients and more effective learning. Another way to regulate the information flow is by introducing additional terms in the loss function to constrain the distribution of membrane potentials~\cite{GuoRecDis-SNN:Networks, Guo2023RMP-Loss:Networks}. Additionally, a number of studies have addressed improving information propagation in deep SNNs by modifying network architectures, incorporating mechanisms such as residual connections and attention to enhance learning and performance~\cite{Hu2024SNNResNet, Shi2024SpikingResformer, Zhou2023Spikingformer}.

While the importance of effective weight initialization for training deep networks is largely acknowledged, weight initialization in SNNs has not received much attention. Some works attempt to empirically determine a suitable weight scale in the case of SNNs, but they often lack a solid theoretical foundation~\cite{Lee2016TrainingBackpropagation, Bellec2018LongNeurons, ZenkeTheNetworks, Herranz-Celotti2022StabilizingTraining}. In~\cite{Ding2022AcceleratingInitialization} the authors derive a new initialization strategy considering the asymptotic spiking response given a mean-driven input. However, this hinders the network from operating in an energy-efficient regime, characterized by very sparse spikes, and undermines the generalization to real-world data, which is typically noisy. \cite{Rossbroich2022Fluctuation-drivenTraining} proposes a fluctuation-driven initialization scheme, but neglects both the spiking and the resetting mechanism, and relies on prior knowledge of the neurons' firing rates for a given dataset. In~\cite{Perez-Nieves2023SpikingCollapse} a specular approach similar to the one presented in \cite{He2015DelvingClassification} is studied, yet the theoretical insights lack empirical validation or comparison with standard initialization methods for ANNs. In this paper, we adopt ~\cite{GlorotUnderstandingNetworks} and~\cite{He2015DelvingClassification} as baselines for ANNs due to their extensive adoption and effectiveness. For SNNs, we considered and implemented the methods introduced in~\cite{Lee2016TrainingBackpropagation}, \cite{Rossbroich2022Fluctuation-drivenTraining}, \cite{Bellec2018LongNeurons},\cite{ZenkeTheNetworks} and \cite{Ding2022AcceleratingInitialization}, as these approaches are specifically tailored to their dynamics.

\section{Methods}
We first introduce the spiking neuron model in Section~\ref{sec:neuron}, then we derive a novel weight initialization method specifically designed for the activation function of a spiking neural network in Section~\ref{sec:derivation}.

\subsection{The spiking neuron}{\label{sec:neuron}}
The Leaky-Integrate-and-Fire (LIF) neuron~\cite{Abbott1999Lapicques1907} is one of the most popular models used in SNNs~\cite{Hunsberger2015SpikingNeurons,Eshraghian2021TrainingLearning} and neuromorphic hardware~\cite{Akopyan2015TrueNorth:Chip,Davies2018Loihi:Learning} to emulate the functionality of biological neurons. The state of a LIF neuron at time $t$ is given by the membrane potential $U(t)$ which evolves according to
\begin{equation}
    \tau \frac{dU(t)}{dt} = -U(t) + RI(t),
\end{equation}
where $\tau$ is the membrane time constant, $R$ is the resistance of the membrane and $I(t)$ is the time-varying input current. Following previous works~\cite{Kim2020RevisitingScratch} we convert the continuous dynamic equation into a discrete equation using the Euler method with discretization time step $\Delta t$. We can represent the membrane potential $u$ at time step $t$ as:
\begin{equation}
    u^{t} = \beta u^{t-1} + \sum\limits_{j}w_{j}x_j^{t},
    \label{eq:snn_discrete}
\end{equation}
where $\beta \propto (1-\Delta t/\tau)$ is a leak factor $\in [0,1]$ governing the rate at which the membrane potential decays over time,  $j$ is the index of the pre-synaptic neuron, $w_j$ represents the connection weights between the pre- and post-synaptic neurons and $x_j$ is the binary spike activation. When the membrane potential $u^t$ exceeds the firing threshold $\theta$, the neuron emits an output spike $x^t=1$. After firing, the membrane potential is reset by subtracting from its value the threshold $\theta$. This operation is conventionally called a `soft reset', in contrast to a `hard reset' (resetting $u$ to 0 after a spike), as it minimizes information loss and can achieve better performance~\cite{GuoReducingNetworks}.

\subsection{Proposed SNN initialization method}\label{sec:derivation}
Our derivation is inspired by He et al.~\cite{He2015DelvingClassification}, which suggests that an effective weight initialization should enable information flow across many network layers by keeping the variance of the input to each layer constant. Similar to the Kaiming initialization, our derivations can be extended to arbitrary feedforward architectures (as shown in Section~\ref{sec:training}). In our analysis here, we focus on the variance of responses within each layer of a fully-connected SNN initialized at time step ${t=0}$. 
For a generic layer $l$ with $m$ neurons:
\begin{align}
    \label{eq:mempot_input}
    \boldsymbol{u}_l &= \boldsymbol{w}_l \boldsymbol{x}_l \\
    \boldsymbol{x}_l &= f(\boldsymbol{u}_{l-1})
\end{align}
Here $\boldsymbol{x}_l \in \{0,1\}^n$ is a binary vector representing the $n$ input spikes, $\boldsymbol{w}_l \in \mathbb{R}^{m \times n}$ is the weight matrix and $\boldsymbol{u}_l \in \mathbb{R}^m$ represents the membrane potentials of neurons in layer $l$. $\boldsymbol{x}_l$ is obtained by applying the activation function $f$ to the membrane potentials of layer $l-1$. In a conventional SNN $f$ is defined as the Heaviside step function:
\begin{eqnarray}
    f({u}_{l-1}) = \begin{cases}
    1, &  \textrm{if } \,\, u_{l-1} > \theta \\
    0, &  \textrm{if } \,\, u_{l-1} < \theta
\end{cases}
\label{eq:activation_function}
\end{eqnarray}
where ${u}_{l-1}$ are the elements of $\boldsymbol{u}_{l-1}$ drawn from the distribution ${P}_{u_{l-1}}$ and $\theta > 0$ is the neuron's firing threshold.
We assume that at initialization the elements of $\boldsymbol{w}_l$ are independently drawn from the same distribution ${P}_{w_l}$ (i.i.d.). Following~\cite{GlorotUnderstandingNetworks} and~\cite{He2015DelvingClassification}, the elements of $\boldsymbol{x}_l$ are also independently drawn from the same distribution ${P}_{x_l}$ (i.i.d.). Lastly, $\boldsymbol{w}_l$ and $\boldsymbol{x}_l$ are independent of each other. We can then write:
\begin{equation}
    \text{Var}[u_{l}] = n_l \text{Var}[w_{l}x_{l}].
\end{equation}
Here ${u}_{l}$, ${w}_{l}$, and ${x}_{l}$ denote the random variables from which the elements of $\mathbf{u}_l$, $\mathbf{w}_l$ and $\mathbf{x}_l$ are respectively sampled. We choose ${w}_{l}$ to be symmetrically distributed around 0. Since ${w}_{l}$ and ${x}_{l}$ are independent of each other, we can rewrite the variance of their product as:
\begin{equation}
    \text{Var}[u_{l}] = n_l \text{Var}[w_{l}]E[x_{l}^2],
    \label{eq:var(u)}
\end{equation}
where $E[x_{l}^2]$ is the expected value of $x_l^2$.
It is worth noting that the expression $E[x_{l}^2]$ strongly depends on the network activation function. Here is where our derivations crucially differ from He et al.~\cite{He2015DelvingClassification}.

\begin{figure}[htbp]
\hspace{-0.15cm}
\centerline{\includegraphics[width=0.48\textwidth, height=4cm]{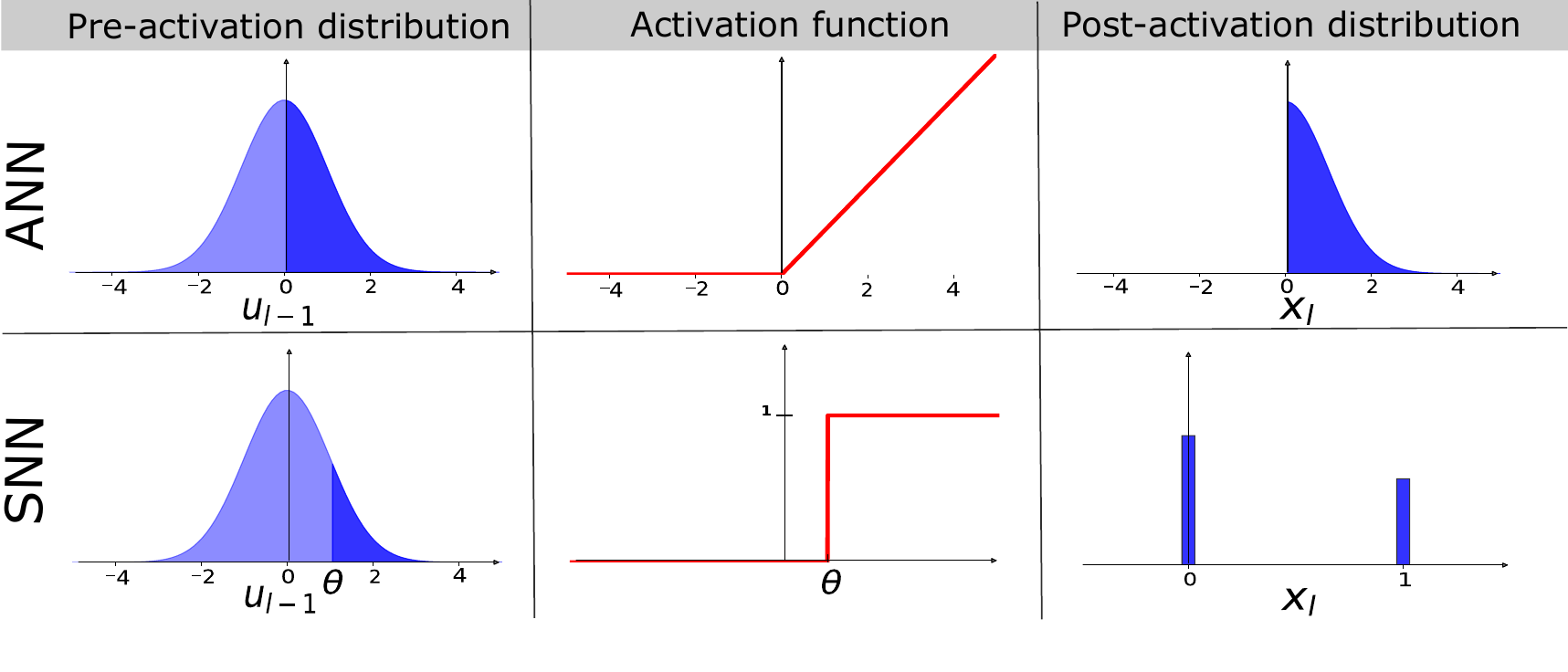}}
\caption{ \textbf{Comparison of standard activation functions for ANNs (\textit{top}) and SNNs (\textit{bottom})}: $\theta$ is the neuron firing threshold. When applied to the pre-activation distribution $u_{l-1}$ (\textit{left}) the SNN thresholding mechanism (\textit{middle}) generates binarized activations $x_l$ (\textit{right}). The dark shaded areas of $u_{l-1}$ correspond to the fraction of neurons which will be activated and provide non-zero input to the next layer. With identical input distributions, this fraction is considerably lower for SNNs. This highlights why weight initializations optimized for ReLU will lead to vanishing activity in deep SNNs.}
\label{fig:activations}
\end{figure}

Assuming $u_{l-1}$ is zero-centered and symmetric around its mean, for the ReLU activation function, ${{x}_{l} = \max(0, u_{l-1})}$, one obtains $E[x_{l}^2] = \frac{1}{2}\text{Var}[u_{l-1}]$. 
This result stems from the fact that the ReLU function preserves exactly the positive half of the distribution it acts upon. As depicted in Figure \ref{fig:activations}, this doesn't hold true for the activation function of SNNs where, by definition, $\theta > 0$. This difference leads to considerably sparser activations in SNNs and to a different conclusion about optimal weight initializations from \cite{He2015DelvingClassification}.
In the case of an SNN, we can express $E[x_{l}^2]$ as:
\begin{equation}
    E[x_{l}^2] = \sum_{j=1}^{n} x_l^j{}^2 P(x_{l} = x_{l}^j).
    \label{eq:E[x^2]}
\end{equation}
Again, the binary elements ${x}_{l}^j \in \{0,1\}$ represent spikes. Applying the SNN activation function \eqref{eq:activation_function} to Eq.~\ref{eq:E[x^2]}, we find that $E[x_{l}^2] = P(u_{l-1} > \theta)$. Equation \ref{eq:var(u)} can then be rewritten as:
\begin{equation}
    \label{eq:varu_withp}
    \text{Var}[u_{l}] = n_l \text{Var}[w_{l}]P(u_{l-1} > \theta).
\end{equation}
As commonly done in recent works~\cite{Guo2023RMP-Loss:Networks,Rathi2020DIET-SNN:Networks,Fang2021DeepNetworks,Fang2021IncorporatingNetworks}, we consider a real-valued input $I_0$ encoded to binary spikes using the first layer of the SNN. When feeding the input to the membrane potentials $u_0$ of the initial layer, $u_0=I_0$ and ${u}_{0}$ trivially follows the same distribution as the input. 
We let $I_0$ be standard normal distributed $I_0 \sim \mathcal{N}(\mu=0, \sigma^2=1)$, thus $\text{Var}[{u}_{0}]~=~1$, $E[u_{0}]~=~0$. 
A proper initialization method should avoid reducing or amplifying the magnitudes of the input signals when propagated across the network layers. In other words, the objective is to prevent vanishing or exploding spikes. This condition can be met if $\text{Var}[u_{l}] = 1$ for every layer $l$, which lets us simplify Eq.~\ref{eq:varu_withp} and leads to a zero-mean Gaussian weight distribution with variance:
\begin{equation}
    \label{eq:optimal_w}
    \text{Var}[w_{l}] = \frac{1}{n_l P(u_{l-1} > \theta)}
\end{equation}
Equation~\ref{eq:optimal_w} is our proposed weight initialization method for training deep SNNs. Note that in terms of architecture parameters, it only depends on the number of input neurons $n$. This makes our initialization applicable to arbitrary architectures and datasets, without any prior knowledge other than $n$.
Because ${u}_{l-1}$ is symmetric around 0 and $\theta > 0$, then $P(u_{l-1} > \theta) < \frac{1}{2}$. It is therefore important to note that:
\begin{equation}
    \frac{1}{n_l P(u_{l-1} > \theta)} > \frac{2}{n_l},
\end{equation}
where $\frac{2}{n_l}$is the standard initialization for a ReLU network~\cite{He2015DelvingClassification}. Thus, initializing the weights of an SNN using a method designed for conventional ANNs with ReLU activation functions does not ensure the propagation of information from the input throughout the network.

\section{Empirical validation}\label{sec:emp_valid}
In this section, we empirically validate our theoretical findings using numerical simulations of deep SNNs. In Section~\ref{sec:theory}, we show that the simulations agree with our theoretical derivation and demonstrate that our weight initialization is able to effectively propagate activity throughout a deep SNN. Conversely, other initialization methods, not specifically tailored to the SNN activation function, do not achieve similar results, but often lead to vanishing spikes. We also discuss how parameters like firing threshold and layer width affect our theoretical findings, and we highlight the presence of finite-size effects in SNNs.
In Section~\ref{sec:theory_time} we expand our analysis to include multi-time-step simulations. We show that our new initialization strategy maintains its ability to conserve the level of activity across layers and time steps, and we explore variations in neuron hyperparameters which uphold this property.

\subsection{Validation in numerical simulations}{\label{sec:theory}}
For the following analysis, unless otherwise specified, we consider fully-connected SNNs with 100 layers and $n=1000$ LIF neurons in each layer. The input $I_0$ is real-valued and randomly drawn from $\mathcal{N}(\mu=0, \sigma^2=1)$.
Consistent with the derivation in Section~\ref{sec:derivation}, we encode the inputs to binary spikes by feeding them to the membrane potentials of the initial LIF layer ${u}_{0}$.

\begin{figure}[htbp]
\hspace{-0.25cm}
\centerline{\includegraphics[width=0.52\textwidth, height=3.45cm]{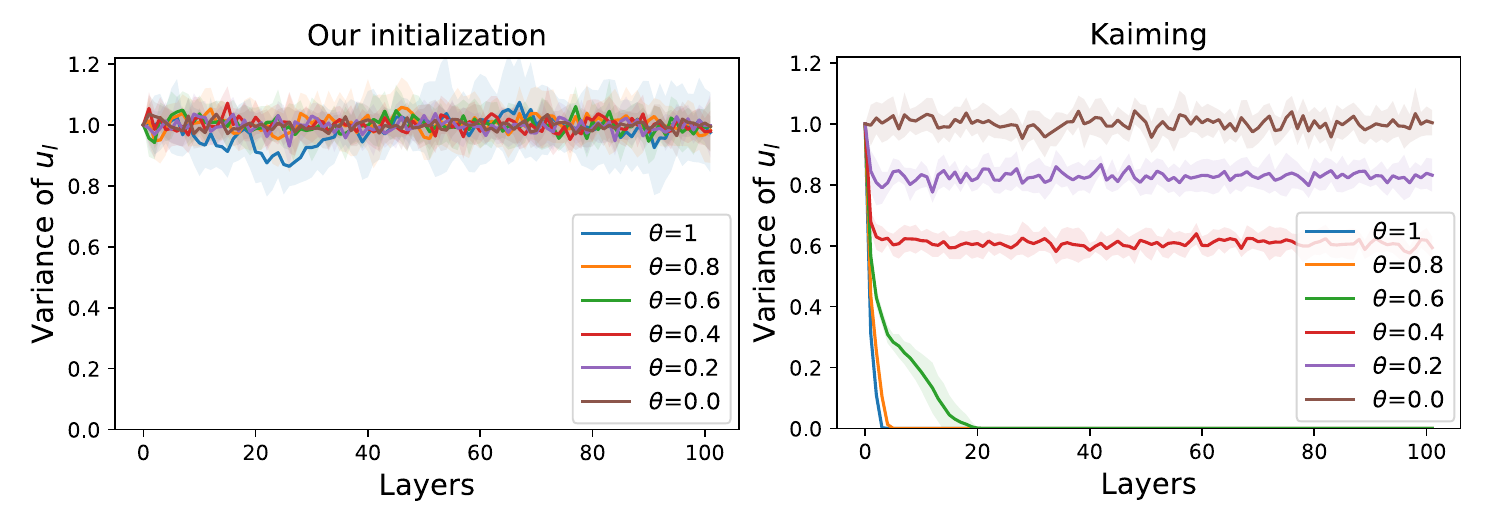}}
\caption{\textbf{Propagation of $\bm{\text{Var}[u_l]}$ across network layers for (\textit{left}) our initialization scheme and (\textit{right}) Kaiming for six firing threshold values($\bm{\theta}$)}: for all $\theta$, our proposed initialization method enables information propagation across all 100 layers. In contrast, Kaiming initialization leads to information dissipation across layers, particularly evident with threshold values close to the standard ${\theta=1}$. Each simulation was repeated 20 times, and the shaded areas represent the standard deviation over these runs.}
\label{fig:theory_theta}
\end{figure}

We investigate the behavior of activity propagation under different weight initialization schemes. First, we compare our method against the prevailing choice for conventional ANNs: Kaiming initialization~\cite{He2015DelvingClassification}.
The weights in the network are therefore randomly initialized respectively from $\mathcal{N}(0, \sqrt{\frac{1}{n P(u_{0} > \theta)}})$ (our method) and $\mathcal{N}(0, \sqrt{\frac{2}{n}})$ (Kaiming), where $n$ is the layer width. Since $u_{0} \sim \mathcal{N}(0,1)$,  $P(u_{0} > \theta)$ is defined as:
\begin{equation}\label{eq:integral}
     P(u_{0} > \theta) = \int_{\theta}^{\infty} \frac{1}{\sqrt{2\pi}} e^{-\frac{u_{0}^2}{2}} du_{0}.
\end{equation}
The integral in Eq.~\ref{eq:integral} doesn't have a closed-form solution, but it can be numerically estimated using the error function~\cite{Andrews1998SpecialEngineers}.
The purpose of our initialization is to retain spiking activity over depth, by conserving $\text{Var}[u_l]$ across the layers. 
To investigate the change in this variance across depth, we initialize the network at time $t=0$, propagate the input across its layers, record the values of the membrane potentials $u_l$ in every layer $l$ and compute their variance.
Figure~\ref{fig:theory_theta} shows how $\text{Var}[u_l]$ evolves with depth for the 2 different initialization schemes and for 6 different values of the firing threshold $\theta$. For every different value of $\theta$, we run the simulation 20 times and plot the average. The shaded areas represent the standard deviation over the different runs.

The results demonstrate that in an SNN initialized with our proposed method (Figure~\ref{fig:theory_theta}, left), the variance $\text{Var}[u_l]$ of the neuron states stays constant across layers as the theory predicts. Specifically, we show that neuronal activity seamlessly propagates across all 100 layers, regardless of the threshold value $\theta$. Conversely, when the network weights are initialized using Kaiming (Figure~\ref{fig:theory_theta}, right), information dissipates across the layers. To alleviate this effect, it is necessary to decrease the standard threshold value ${\theta=1}$, to enable more spikes. Notably, using Kaiming, the only viable method to effectively preserve information is by setting ${\theta=0}$, where the activation function becomes effectively equivalent to ReLU (Figure~\ref{fig:activations}, left).

\begin{figure}[htbp]
\hspace{-0.35cm}
\centerline{\includegraphics[width=0.53\textwidth, height=3.28cm]{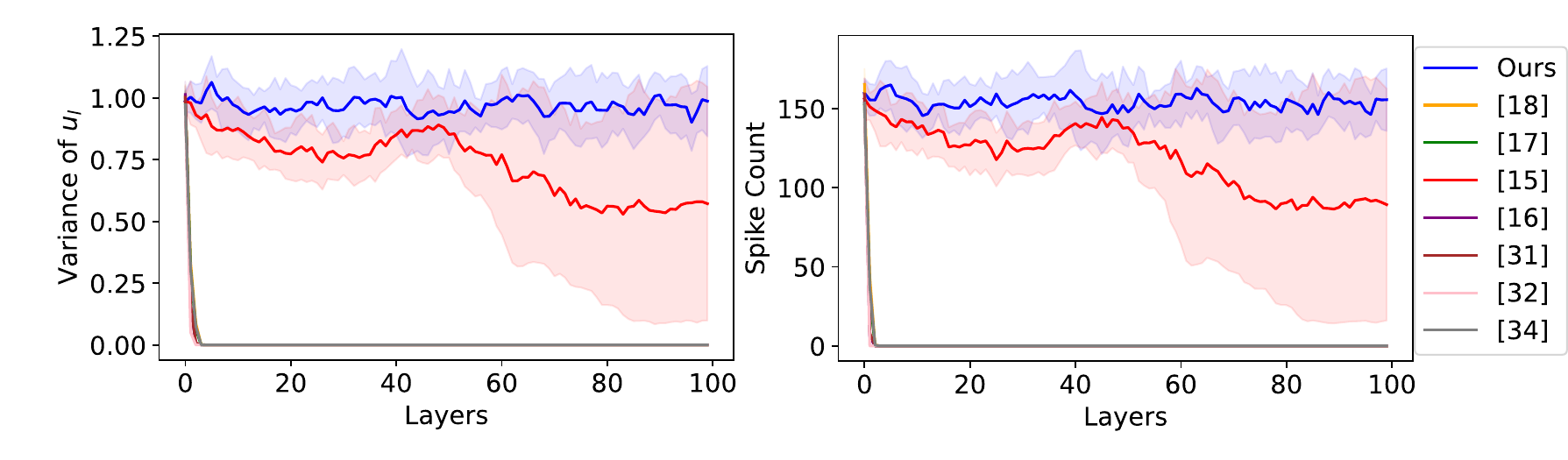}}
\caption{\textbf{Propagation of (\textit{left}) $\bm{\text{Var}[u_l]}$ and (\textit{right}) total number of spikes across network layers for our method and baseline approaches from the literature}: comparing our method (blue line) with different initialization schemes from the literature for both ANNs and SNNs for $\theta=1$. Our proposed initialization method maintains a constant variance $\text{Var}[u_l]$, therefore enabling the propagation of spikes across all 100 layers. In contrast, other methods struggle to conserve activity. Each simulation was repeated 20 times, and the shaded areas represent the standard deviation over these runs.}
\label{fig:1ts_baselines}
\end{figure}

Additionally, we repeat the analysis described above, comparing our method with seven other weight initialization schemes proposed in the literature for both ANNs~(\cite{GlorotUnderstandingNetworks}, \cite{He2015DelvingClassification}) and SNNs~(\cite{Lee2016TrainingBackpropagation, Bellec2018LongNeurons, ZenkeTheNetworks}, \cite{Ding2022AcceleratingInitialization}, \cite{Rossbroich2022Fluctuation-drivenTraining}). For this analysis, we set $\theta=1$ and record the variance $\text{Var}[u_l]$ and the total number of spikes emitted per layer. As shown in Figure~\ref{fig:1ts_baselines}, preserving the variance directly corresponds to maintaining effective spike propagation. The results indicate that nearly all alternative initialization methods, except for\cite{Rossbroich2022Fluctuation-drivenTraining}, fail to sustain activity beyond approximately 5 layers.\\

\begin{figure}[htbp]
\hspace{-0.4cm}
\centerline{\includegraphics[width=0.52\textwidth, height=3.45cm]{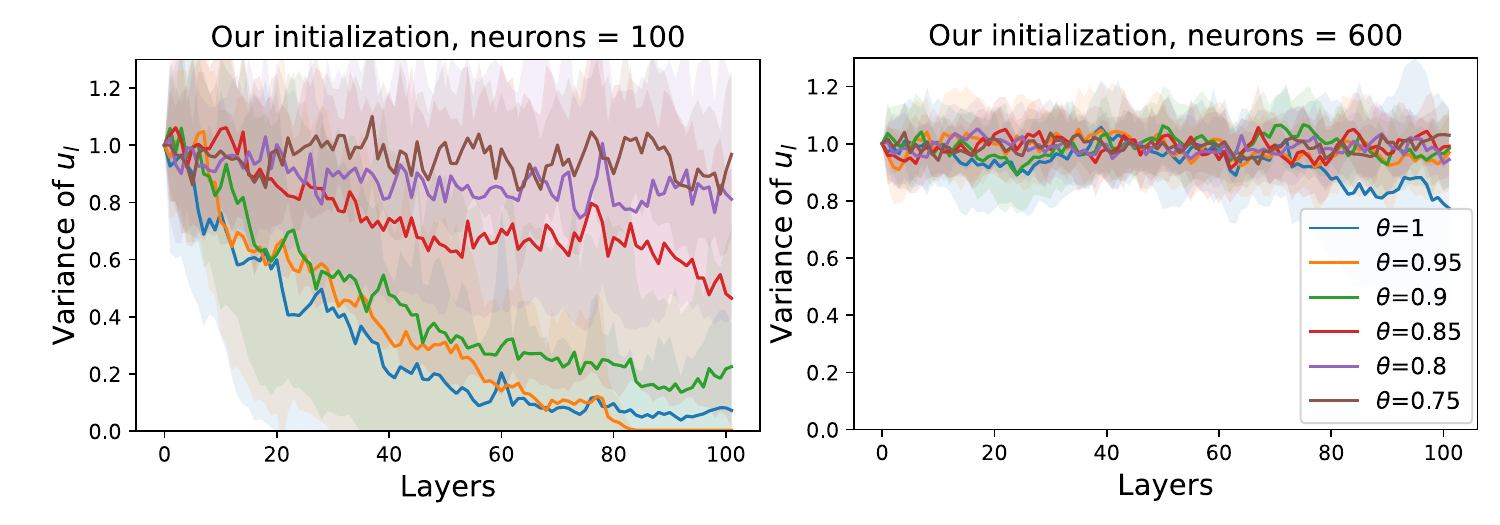}}
\caption{\textbf{Impact of the finite-size effect on the propagation of $\bm{\text{Var}[u_l]}$ across layers}: as the number of neurons decreases and spiking threshold $\theta$ increases,  the discrepancy between empirical and theoretical results becomes more pronounced. (\textit{Left}): $n = 100$. The network can't conserve activity over depth for $\theta>0.85$. (\textit{Right}): $n=600$. By increasing the number of neurons the finite-size effect becomes less pronounced even for higher values of $\theta$. Each simulation was repeated 20 times, and the shaded areas represent the standard deviation over these runs.}
\label{fig:theory_finite_size}
\end{figure}

We then explore the scalability of our approach with respect to the width of the layers.
First, we note that, the higher the firing threshold, the smaller the area defined by the integral in Eq.~\ref{eq:integral}. This means that the probability of spiking is decreased. For high $\theta$, where spiking probability is theoretically low, as we reduce the number of neurons $n$ in each layer, it becomes increasingly less likely to numerically sample the estimated `average' number of spikes in the simulations. Specifically, since we cannot sample the necessary number of spikes in layers with small $n$, the activity will die out in deeper layers. This deviation of simulations from theory does not happen if the layer size $n$ is large enough compared to the spiking probability of Eq.~\ref{eq:integral}. We refer to this phenomenon as a finite-size effect, and its impact on the propagation of information in the network is illustrated in Figure~\ref{fig:theory_finite_size}.

The results demonstrate that a 100-layer network with layer width $n=100$ struggles to retain the spiking activity in deeper layers as we increase $\theta$ (Figure~\ref{fig:theory_finite_size}, left). The finite-size effect becomes particularly prominent for values of $\theta$ exceeding 0.85. However, when we increase the layer width from 100 to 600, we observe a significant reduction in the prominence of this effect, except for the case of $\theta = 1$ (Figure~\ref{fig:theory_finite_size}, right).
We find this insight particularly relevant in the context of designing SNN architectures and determining their hyperparameters. Due to the thresholding operation serving as the activation function in an SNN and the binarization of activations, the combination of the number of neurons and firing threshold significantly influences the network's behavior. Specific configurations of these parameters can in fact complicate information propagation if proper adjustments are not implemented. Finally, we note that our theory holds exactly in simulations with threshold $0 \geq \theta \geq 1$ and $n > 600$.

\subsection{Extension to multiple time steps}{\label{sec:theory_time}}
SNNs demonstrate their main advantages when processing time-dependent input. This is because spiking neurons, inspired by biological counterparts, possess an intrinsic memory, the membrane potential $u$, which naturally integrates information over time. Through precise spike timing, SNNs can process temporal information with greater computational power than ANNs~\cite{Maass1997NetworksModels}. This makes them well-suited for tasks involving dynamic temporal patterns, such as speech recognition and video analysis \cite{Pellegrini2020Low-activityRecognition,Wu2020DeepRecognition,Liu2021Event-basedNetworks}.

In this section we extend the analysis of Section~\ref{sec:theory} to activity propagation in space \textit{and} time. We address the question of how weight initialization affects information propagation in multiple time-step simulations of deep SNNs.
We employ fully-connected SNNs of 100 layers with $n=1000$ neurons in each layer. We consider LIF neurons with soft reset and numerically compute the discrete-time dynamics based on Eq.~\ref{eq:snn_discrete}. The dynamics of the membrane potentials including the reset term is given by:
\begin{equation} 
    \mathbf{u}_l^{t} = \mathbf{w}_l^{t} \mathbf{x}_l^{t} + \beta \mathbf{u}_{l}^{t-1} - \mathbf{x}_{l+1}^{t-1} \theta,
    \label{eq:reset}
\end{equation}
for time step $t > 0$ and layer $l$. $\beta \in [0, 1]$ is again the leak factor. Inputs to the first layer are randomly drawn from ${\sim\mathcal{N}(0,1)}$ and we set $\theta=1$ and $\beta=0.5$, same as in Section~\ref{sec:theory} (Figure~\ref{fig:1ts_baselines}).
Differently, in this section, we iteratively feed the input (constant over time) to the membrane potentials of the initial LIF layer ${u}_{0}^{t}$ at every time step $t$.
As in Section~\ref{sec:theory}, we first compare our initialization method against Kaiming initialization. We compute the variance of the membrane potentials ${u}_{l}^{t}$ and the total spike count at each layer $l$ and time step $t$, for a total of $T=20$ time steps.

\begin{figure}[htbp]
\vspace{-0.1cm}
\centerline{\includegraphics[width=0.42\textwidth, height=6.7cm]{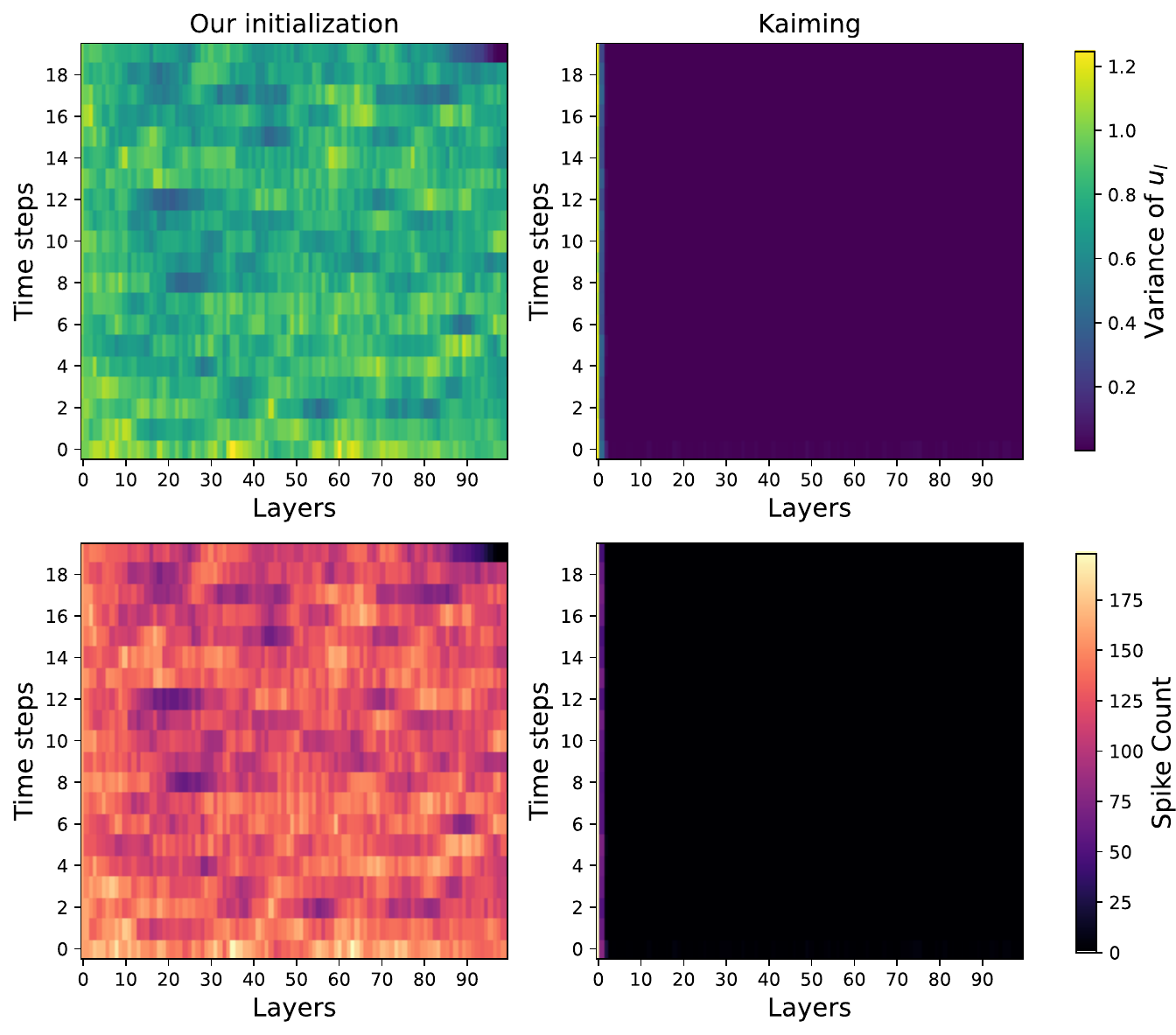}}
\caption{ \textbf{Propagation of $\text{Var}[\bm{u}_{l}^{t}]$ (\textit{top row}) and number of spikes (\textit{bottom row}) across layers and time steps for our initialization method (\textit{left}) and Kaiming (\textit{right})}: our proposed weight initialization preserves activity and propagates spikes through 100 layers and 20 time steps. In contrast, with Kaiming initialization neuronal activity dies out after a few layers.}
\label{fig:multisteps}
\end{figure}

The network initialized with our method succeeds in conserving $\text{Var}[u_{l}^{t}]$ across both space and time, whereas the network initialized with Kaiming dramatically fails at it (Fig.~\ref{fig:multisteps}).

We then repeat the analysis using our initialization scheme and the strategies proposed in~\cite{GlorotUnderstandingNetworks, He2015DelvingClassification}, \cite{Lee2016TrainingBackpropagation, Bellec2018LongNeurons, ZenkeTheNetworks}, \cite{Ding2022AcceleratingInitialization} and~\cite{Rossbroich2022Fluctuation-drivenTraining}.
For each initialization method, we compute the variance of the membrane potentials ${u}_{l}^{t}$ and the total spike count at each layer $l$ and time step $t$, for a total of $T=20$ time steps. The results are then avaraged across the time dimension. Each simulation is repeated 10 times, and we report the mean of these runs.

\begin{figure}[htbp]
\hspace{-0.3cm}
\centerline{\includegraphics[width=0.525\textwidth, height=3.27cm]{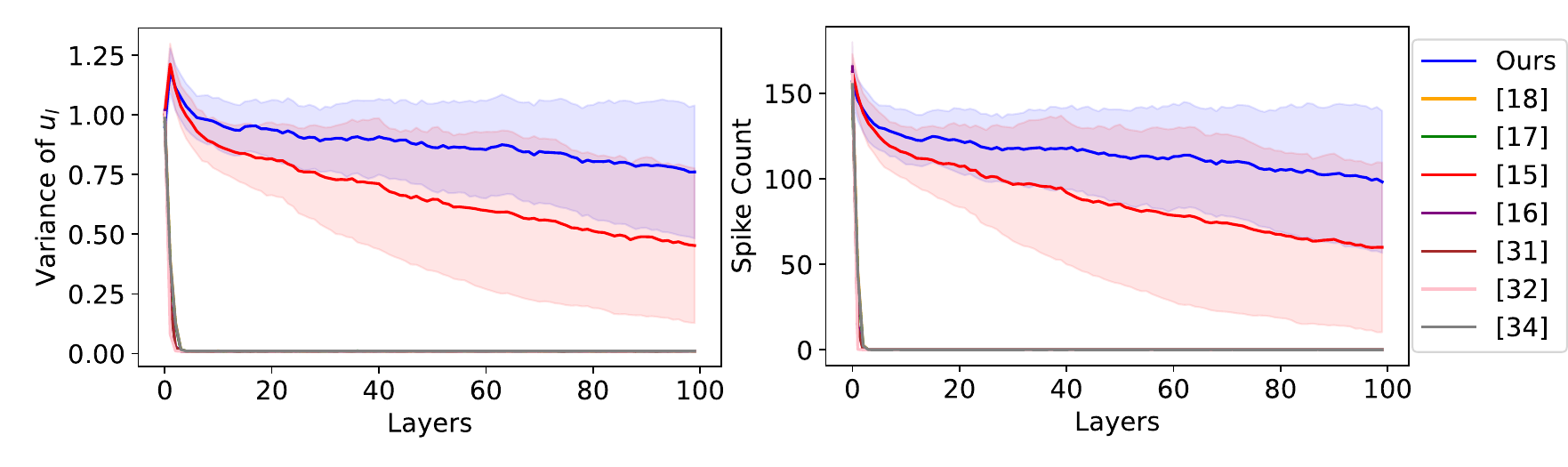}}
\caption{ \textbf{Propagation of (\textit{left}) $\text{Var}[u_l]$ and (\textit{right}) total number of spikes across network layers averaged across $\bm{T=20}$ time steps}: we compare our method (blue line) with other initialization schemes from the literature for both ANNs and SNNs. While our method does not explicitly account for time, it approximately conserves variance and provides consistent network output, in contrast to the other methods. Each simulation was repeated 10 times, with shaded areas indicating the standard deviation across these runs.}
\label{fig:baselines}
\end{figure}

We observe a slow decrease in the spike count, but our initialization method succeeds in conserving $\text{Var}[u_{l}^{t}] $ across both space and time relatively well compared to other methods (Fig.~\ref{fig:baselines}). Conserving  $\text{Var}[u_{l}^{t}]$ is crucial, as it means conserving activity, and therefore ensuring a consistent network output.

Although our mathematical derivation does not explicitly take time into account, it's an improvement on using ANN initializations as we consider the specific SNN activation function. By keeping the variance of the membrane potentials $u_l$ constant, our method indirectly constrains the network to keep the variance of the layer input $w_lx_l$ constant (Eq.~\ref{eq:mempot_input}). This helps to effectively propagate information also across multiple time steps. Nevertheless, we expect deviations from theory, caused by the leak and reset terms (Eq.~\ref{eq:reset}).

\begin{figure}[htbp]
\hspace{-0.2cm}
\centerline{\includegraphics[width=0.53\textwidth, height=3.23cm]{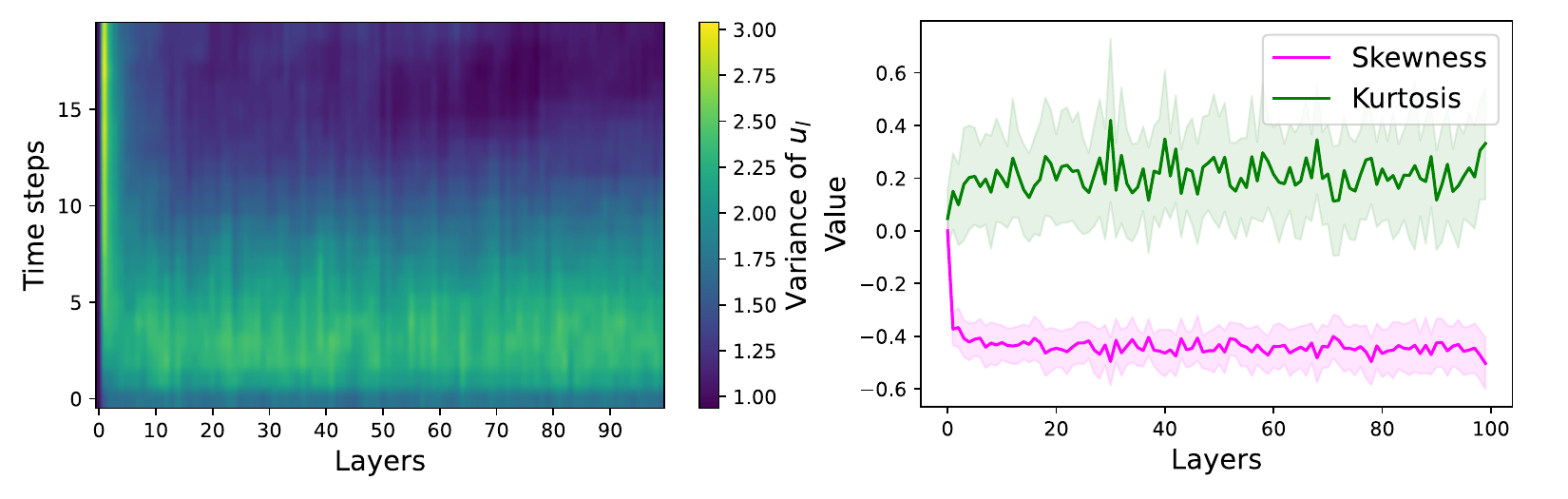}}
\caption{\textbf{For $\bm{\beta=0.9}$: (\textit{left}) propagation of $\bm{\text{Var}}[u_l]$ across network layers over $\bm{T=20}$ time steps, (\textit{right}) skewness and excess kurtosis of $\bm{u_l^t}$ across layers, averaged over $\bm{T=20}$ time steps}: for high $\beta$, the reset operation violates the assumption that ${u}_{l}^{t}$ is always normally distributed and symmetrically centered around 0. This effect is evident in the reduced stability of $\text{Var}[u_l]$ across layers and time steps, and in the non-zero values of skewness and kurtosis. Each simulation was repeated 10 times and the shaded areas indicate the standard deviation across the runs. }
\label{fig:kurtosis}
\end{figure}

In particular, the reset operation has a non-negligible effect on the distribution of the membrane potentials. This operation violates the assumption that ${u}_{l}^{t}$ is always normally distributed and symmetrically centered around 0. However, how well the normal distribution still holds as an approximation depends on neuron hyperparamters. For example, deviations from theory are visible at higher values of $\beta$. A larger $\beta$ leads to broader distributions of ${u}_{l}^{t}$, and thus to a more abrupt change in the distributions when neurons with ${u}_{l}^{t} > \theta $ are reset.
As illustrated in Fig.~\ref{fig:kurtosis} (left), when $\beta = 0.9$, the $\text{Var}[u_l]$ exhibits less stability across layers and time steps. We attribute this instability to the shift in ${u}_{l}^{t}$ distributions. In fact, in Figure  \ref{fig:kurtosis} (right) we present the skewness and excess kurtosis values for \textit{$u_l^t$} across layers and time steps. Skewness measures the degree of asymmetry of the distribution, while excess kurtosis measures the degree of peakedness and flatness of a distribution. For reference, a normal distribution has 0 skewness and 0 excess kurtosis. We observe that, in this case, \textit{$u_l^t$} tends
to a left-skewed and heavy-tailed distribution, thus diverging from a normal distribution. Nevertheless, despite this deviation from the analytical solution, with the proposed method we're still able to retrieve a consistent network output.

\section{Evaluation on classification datasets}{\label{sec:training}}

\begin{figure*}[htbp]
\centerline{\includegraphics[width=0.8\textwidth, height=4cm]{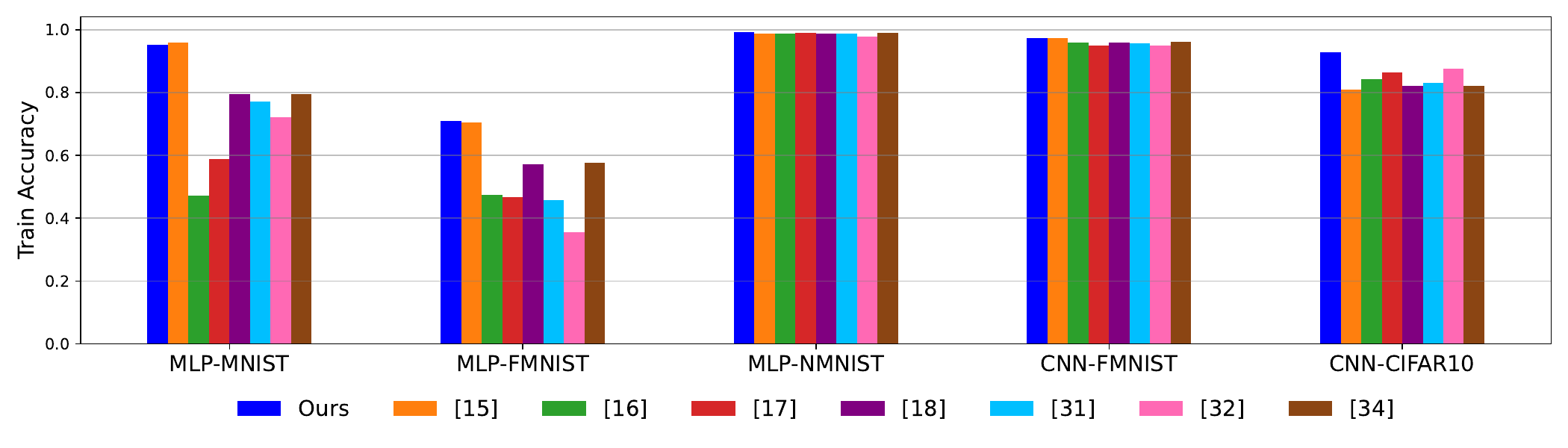}}
\caption{\textbf{Training accuracy across 4 image classification datasets and 2 network architectures}: comparing our weight initialization method against 7 methods from the literature for both ANNs and SNNs. For the MLP, only our method (blue) and~\cite{Rossbroich2022Fluctuation-drivenTraining} (orange) achieve over $90\%$ accuracies on MNIST and $70\%$ on FMNIST, with our approach showing a slight advantage on FMNIST. All methods perform similarly on NMNIST. For the CNN, all methods perform well on FMNIST, with ours and~\cite{Rossbroich2022Fluctuation-drivenTraining} show marginally better results. Notably, our approach significantly outperforms the others on CIFAR-10. }
\label{fig:training}
\end{figure*}

In this section, we evaluate how our variance-conserving weight initialization for SNNs can translate into accelerated training, improved accuracy and lower latency. To that end, we conduct image classification experiments using both fully-connected and convolutional SNNs. 
Unless otherwise specified, our fully-connected SNN (MLP) consist of 10 layers, each comprising $n=600$ LIF neurons with soft reset, ${\theta=1}$ and ${\beta=0.5}$.
The convolutional SNN (CNN) consists of six convolutional layers, each providing input to a layer of LIF neurons with ${\beta=0.5}$ and ${\theta=1}$. The final layer is a fully connected linear layer. Each convolutional layer has 64 channels, $3\times 3$ kernels, and padding of 1, preserving spatial dimensions across layers. For a convolutional layer \( l \), the \( n_l \) in Eq. \ref{eq:optimal_w} is computed as $n_{l} = k^2 c$, where \( k*k\) is the kernel size and \( c \) is the number of input channels.
For input encoding, the MLP converts inputs to binary spikes through the first LIF layer, while the CNN directly passes the input to the first convolutional layer. In both architectures the final layer of the network outputs binary spikes, which are accumulated over time steps and passed to the cross-entropy loss function. The cross entropy loss encourages the correct class to fire at all
time steps, and aims to suppress incorrect classes from firing. We considered four different datasets: MNIST~\cite{Lecun1998Gradient-basedRecognition}, Fashion MNIST (FMNIST)~\cite{xiao2017fashionmnist}, CIFAR-10~\cite{krizhevsky2009learning}, and Neuromorphic MNIST (NMNIST)~\cite{orchard2015converting}.
The MNIST and FMNIST datasets each consist of 60,000 training samples and 10,000 test samples, where each sample is a 28×28 pixel grayscale image representing one of ten classes. CIFAR-10 consists of 60,000 color images (32x32 pixels), divided into 50,000 training images and 10,000 test images, with each image classified into one of ten categories.
For all datasets, pixel intensities are normalized to have a mean of 0 and variance of 1, following the assumptions in our theoretical derivations.
The NMNIST dataset provides a spiking neuromorphic representation of the original MNIST digits. It includes 60,000 training samples and 10,000 test samples, where each sample is represented as a 28×28 pixel event-based recording of a digit. Events are binned into frames using a time window of 1000\ $\mu\text{s}$.
 
Commonly, SNNs performing spike-count based object classification on standard image datasets require a large number of total time steps $T$, in order to accumulate a reliable number of spikes at the output layer. A typical range of $T$ can be between 10 and a few thousand~\cite{Roy2019TowardsComputing}. Here, in order to demonstrate the power of a good weight initialization, we set the number of total time steps to $T=3$. We hypothesize that initializations which enable constant information propagation across depth might also enable inference with low latency, where there is no need to wait for many time steps to accumulate the necessary number of output spikes.

The networks are trained using backpropagation through time (BPTT) \cite{Lee2016TrainingBackpropagation} and the arctan surrogate gradient function~\cite{Fang2021IncorporatingNetworks}. We utilize the Adam optimizer \cite{Kingma2014Adam:Optimization} with an initial learning rate of ${1 \times 10^{-3}}$ and employ cosine annealing scheduling. Training is conducted for 150 epochs across all datasets, except for CIFAR-10, where the number of epochs is increased to 300. We compare our weight initialization method with other commonly used initialization techniques from the literature for both ANNs and SNNs, as in previous sections.

\begin{figure*}[htbp]
\centerline{\includegraphics[width=0.8\textwidth, height=4cm]{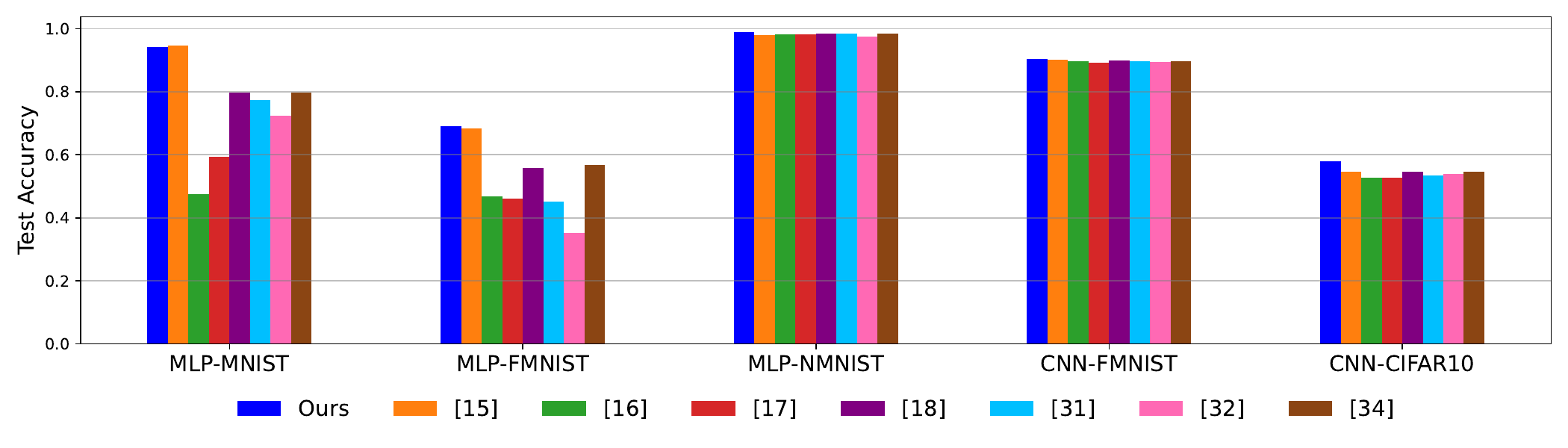}}
\caption{\textbf{Test accuracy across 4 image classification datasets and 2 network architectures}: comparing our weight initialization method against 7 methods from the literature for both ANNs and SNNs. For the MLP, only our method (blue) and~\cite{Rossbroich2022Fluctuation-drivenTraining} (orange) achieve over $90\%$ accuracies on MNIST and $68\%$ FMNIST, with our approach showing a slight advantage on FMNIST. All methods perform similarly on NMNIST. For the CNN, all methods perform well on FMNIST, while our approach achieves better accuracy than the others on CIFAR-10.}
\label{fig:test}
\end{figure*}

The results are shown in Figure~\ref{fig:training} and Figure~\ref{fig:test}. Using the MLP architecture, we find that networks initialized with different methods achieve similar performance on NMNIST, likely due to the inherent information redundancy in this dataset. Specifically, even though the samples in NMNIST are temporal sequences of events that are grouped into frames, all the information necessary to perform the task is encoded primarily in the spatial domain. The temporal information, while present, does not contribute significantly to the task, making it largely redundant for solving the classification problem. However, for MNIST and FMNIST, only our method and that of~\cite{Rossbroich2022Fluctuation-drivenTraining} reach training accuracies above $90\%$ and $70\%$, respectively, after 150 epochs, with our method showing a slight advantage on FMNIST. As observed in Figure \ref{fig:baselines}, these two methods were most effective at preserving information across network layers and time steps, suggesting that this property may contribute to their improved performance.
For the CNN architecture, all initialization methods perform well on FMNIST, but our method outperforms the others on CIFAR-10.

\begin{figure}[htbp]
\centerline{\includegraphics[width=0.535\textwidth, height=5.85cm]{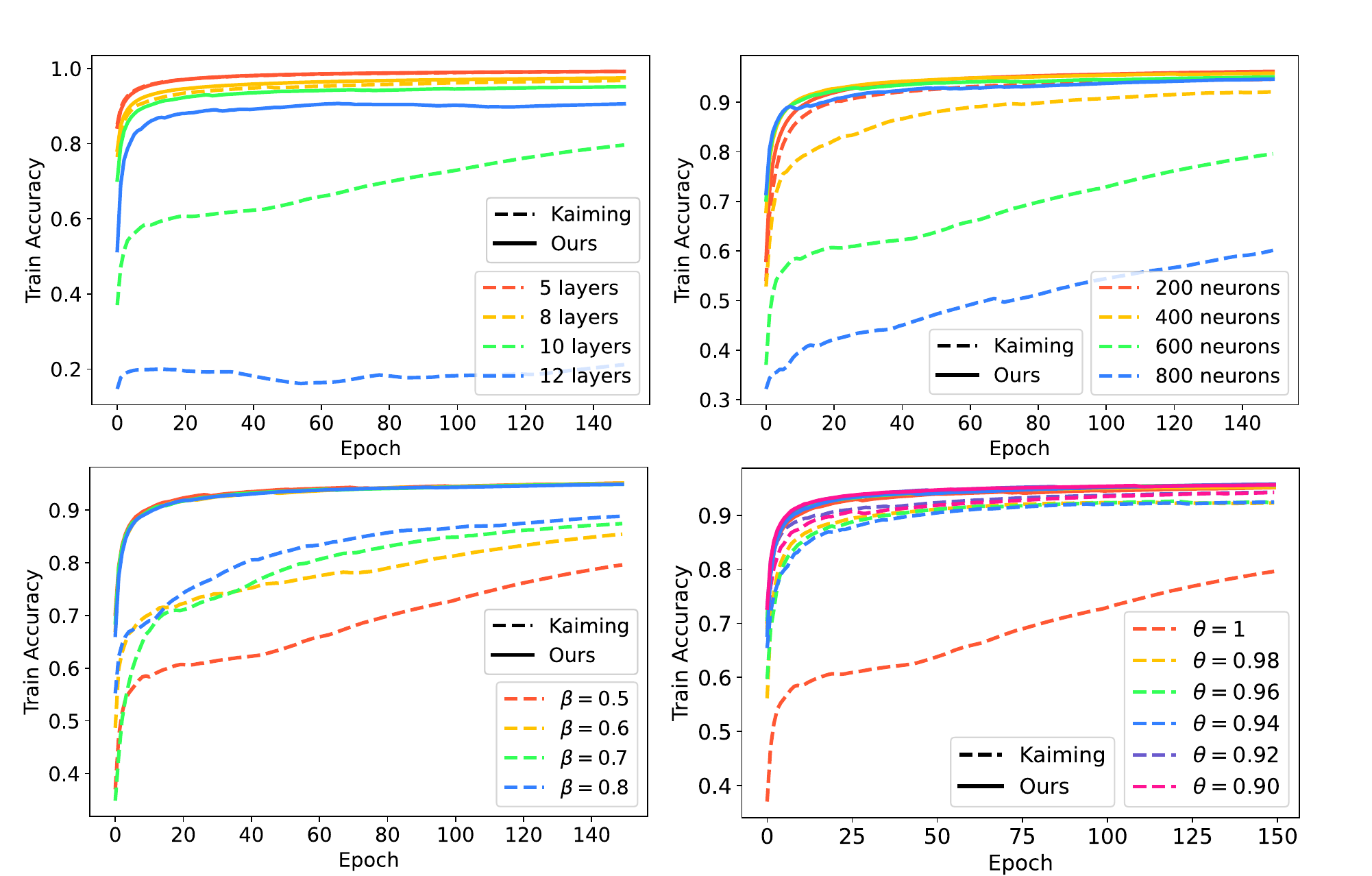}}
\caption{\textbf{Robustness to variations in network and neuron parameters for our method and Kaiming}: training accuracy of an MLP trained on MNIST for different values of (\textit{top left}) network depth, (\textit{top right}) layer width, (\textit{bottom left}) $\beta$ and (\textit{bottom right}) $\theta$. We compare our proposed initialization method (solid lines) to Kaiming (dashed lines) and find that it achieves better training accuracy, faster convergence and it remains robust across different network and neuron parameters.}
\label{fig:mnist_train}
\end{figure}

We then examine the robustness of our weight initialization method under variations in network and neuron parameters using the MLP architecture described above, trained on MNIST. Specifically, we test different values for network depth, layer width, $\beta$, and $\theta$ and find that our initialization method remains stable and effective across these variations. We compare our method with Kaiming initialization, the prevailing choice for conventional ANN, and observe that our approach enables faster convergence during training and results in higher training accuracies. Networks initialized with Kaiming show slower convergence, worse performance and significant variability in behavior across different parameter values, indicating a lack of robustness to such changes. Consistent with both theoretical insights and empirical findings presented in previous sections, we attribute this behavior to inadequate activity propagation. Specifically, we note that training becomes increasingly challenging with deeper networks, higher values of $\theta$ (resulting in fewer neurons emitting spikes) and lower values of $\beta$ (leading to less information retention from the previous time step). In contrast, our proposed initialization scheme proves resilient to changes in these critical hyperparameters.

\section{Conclusion and Discussion}\label{sec:conclusion}
Introduction of new initialization techniques has significantly contributed to the overwhelming success of ANNs in recent years, particularly in enabling deeper architectures. It's evident that proper weight initialization plays a crucial role in achieving optimal accuracy and facilitating fast convergence. In this paper, we address the problem of weight initialization in Spiking Neural Networks (SNNs) and show how the techniques developed for ANNs, such as Kaiming initialization, are inadequate for SNNs. Unlike standard non-linearities for ANNs, the thresholding mechanism in SNNs produces binarized activations. Consequently, combining SNNs with weight initializations optimized for ReLU results in the well-known vanishing spikes problem. This inevitably leads to information loss in deeper layers. Here, we analytically derive a novel weight initialization method which takes into account the specific activation function of SNNs. We empirically demonstrate that, in contrast to Kaiming and other commonly used schemes, our initialization enables spiking activity to propagate from the input to the output layers of deep networks without
being dissipated or amplified, a necessary condition for effective network training.
Similar to the popular Kaiming initialization for ANNs, our weight initialization depends only on the number of input neurons $n$ to a layer. Therefore, our approach is broadly applicable to all deep, spiking network architectures with fixed connectivity maps. Furthermore, we observe that the only baseline SNN initialization method with comparable performance to ours is \cite{Rossbroich2022Fluctuation-drivenTraining}. While \cite{Rossbroich2022Fluctuation-drivenTraining} can retain the variance of the activations over a large number of layers, it needs statistics of the datasets, and spiking rates of the specific architecture obtained from simulations of the untrained network. In this way the method in [15] overfits its initialization to the distribution of the training set, and is not dataset-agnostic. In contrast, our method is generalizable to all datasets, without requiring prior knowledge of dataset statistics.

In addition, we demonstrate that our proposed initialization is robust against variations in several important network and neuron hyperparameters, including network depth, layer width, $\beta$ and $\theta$. This robustness can enable deep activity propagation for diverse models and machine learning tasks. To further improve training in deep SNNs, we highlight that, due to the specific properties of the SNN activation function, the width of the network and the firing threshold might significantly affect information propagation.

A limitation of our approach is that the proposed initialization does not account for the temporal dynamics of the neuron membrane potentials \textit{${u}_{l}$}, specifically the effects of the leakage and the resetting operation. Specifically, after neurons which emit a spike are reset, our assumption that \textit{${u}_{l}$} is normally distributed around zero is violated. We observe that the extent to which the normal distribution remains a valid approximation depends on the neuron hyperparameters. Our initialization strategy, unlike Kaiming, still helps to effectively propagate information, also across multiple time steps. However, the theory can be expanded to explicitly take into account the temporal variations in ${u}_{l}$.
Another assumption of our derivation is that the activations $x_{l}$ are mutually independent, but this condition is typically violated in the case of real-world data, such as images. However, the same assumption underlies the popular ANN initialization methods, known to be effective on real-world datasets. Empirically, in Section~\ref{sec:training} we illustrate how, for an SNN trained on 4 different image classification datasets, our variance-conserving initialization scheme still translates into accelerated training, improved accuracy and low latency, especially when compared to Kaiming.  It is important to note that the 10-layer MLP network was not selected to achieve high accuracy, but rather to benchmark the impact of different initialization methods, which is why its performance may appear suboptimal. We acknowledge the necessity of extending this analysis to more complex architectures and datasets, in order to evaluate its effectiveness in various settings.

\bibliographystyle{unsrt} 
\bibliography{references}

\begin{thebibliography}{10}

\bibitem{Abbott1999Lapicques1907}
L~F Abbott.
\newblock {Lapicque's introduction of the integrate-and-fire model neuron (1907)}, 1999.

\bibitem{Hunsberger2015SpikingNeurons}
Eric Hunsberger and Chris Eliasmith.
\newblock {Spiking Deep Networks with LIF Neurons}, 10 2015.

\bibitem{Maass1997NetworksModels}
Wolfgang Maass.
\newblock {Networks of Spiking Neurons: The Third Generation of Neural Network Models}, 1997.

\bibitem{Indiveri2011NeuromorphicCircuits}
Giacomo Indiveri, Bernabé Linares-Barranco, Tara~Julia Hamilton, André van Schaik, Ralph Etienne-Cummings, Tobi Delbruck, Shih~Chii Liu, Piotr Dudek, Philipp H{\"{a}}fliger, Sylvie Renaud, Johannes Schemmel, Gert Cauwenberghs, John Arthur, Kai Hynna, Fopefolu Folowosele, Sylvain Saighi, Teresa Serrano-Gotarredona, Jayawan Wijekoon, Yingxue Wang, and Kwabena Boahen.
\newblock {Neuromorphic silicon neuron circuits}, 2011.

\bibitem{Maass2004OnNeurons}
Wolfgang Maass and Henry Markram.
\newblock {On the computational power of circuits of spiking neurons}.
\newblock {\em Journal of Computer and System Sciences}, 69(4):593--616, 2004.

\bibitem{Davies2018Loihi:Learning}
Mike Davies, Narayan Srinivasa, Tsung~Han Lin, Gautham Chinya, Yongqiang Cao, Sri~Harsha Choday, Georgios Dimou, Prasad Joshi, Nabil Imam, Shweta Jain, Yuyun Liao, Chit~Kwan Lin, Andrew Lines, Ruokun Liu, Deepak Mathaikutty, Steven McCoy, Arnab Paul, Jonathan Tse, Guruguhanathan Venkataramanan, Yi~Hsin Weng, Andreas Wild, Yoonseok Yang, and Hong Wang.
\newblock {Loihi: A Neuromorphic Manycore Processor with On-Chip Learning}.
\newblock {\em IEEE Micro}, 38(1):82--99, 1 2018.

\bibitem{Akopyan2015TrueNorth:Chip}
Filipp Akopyan, Jun Sawada, Andrew Cassidy, Rodrigo Alvarez-Icaza, John Arthur, Paul Merolla, Nabil Imam, Yutaka Nakamura, Pallab Datta, Gi~Joon Nam, Brian Taba, Michael Beakes, Bernard Brezzo, Jente~B. Kuang, Rajit Manohar, William~P. Risk, Bryan Jackson, and Dharmendra~S. Modha.
\newblock {TrueNorth: Design and Tool Flow of a 65 mW 1 Million Neuron Programmable Neurosynaptic Chip}.
\newblock {\em IEEE Transactions on Computer-Aided Design of Integrated Circuits and Systems}, 34(10):1537--1557, 10 2015.

\bibitem{Yamazaki2022SpikingReview}
Kashu Yamazaki, Viet~Khoa Vo-Ho, Darshan Bulsara, and Ngan Le.
\newblock {Spiking Neural Networks and Their Applications: A Review}, 7 2022.

\bibitem{Deng2021OptimalNetworks}
Shikuang Deng and Shi Gu.
\newblock {Optimal Conversion of Conventional Artificial Neural Networks to Spiking Neural Networks}, 2 2021.

\bibitem{Ding2021OptimalNetworks}
Jianhao Ding, Zhaofei Yu, Yonghong Tian, and Tiejun Huang.
\newblock {Optimal ANN-SNN Conversion for Fast and Accurate Inference in Deep Spiking Neural Networks}, 5 2021.

\bibitem{Ho2020TCL:Layers}
Nguyen-Dong Ho and Ik-Joon Chang.
\newblock {TCL: an ANN-to-SNN Conversion with Trainable Clipping Layers}, 8 2020.

\bibitem{Neftci2019SurrogateNetworks}
Emre~O. Neftci, Hesham Mostafa, and Friedemann Zenke.
\newblock {Surrogate Gradient Learning in Spiking Neural Networks: Bringing the Power of Gradient-based optimization to spiking neural networks}.
\newblock {\em IEEE Signal Processing Magazine}, 36(6):51--63, 11 2019.

\bibitem{Zenke2021TheNetworks}
Friedemann Zenke and Tim~P. Vogels.
\newblock {The Remarkable Robustness of Surrogate Gradient Learning for Instilling Complex Function in Spiking Neural Networks}.
\newblock {\em Neural computation}, 33(4):899--925, 3 2021.

\bibitem{Hochreiter1997LongMemory}
Sepp Hochreiter and Jürgen Schmidhuber.
\newblock {Long Short-Term Memory}.
\newblock {\em Neural Computation}, 9(8):1735--1780, 11 1997.

\bibitem{Rossbroich2022Fluctuation-drivenTraining}
Julian Rossbroich, Julia Gygax, and Friedemann Zenke.
\newblock {Fluctuation-driven initialization for spiking neural network training}, 6 2022.

\bibitem{Lee2016TrainingBackpropagation}
Jun~Haeng Lee, Tobi Delbruck, and Michael Pfeiffer.
\newblock {Training Deep Spiking Neural Networks using Backpropagation}, 8 2016.

\bibitem{GlorotUnderstandingNetworks}
Xavier Glorot and Yoshua Bengio.
\newblock {Understanding the difficulty of training deep feedforward neural networks}.

\bibitem{He2015DelvingClassification}
Kaiming He, Xiangyu Zhang, Shaoqing Ren, and Jian Sun.
\newblock {Delving Deep into Rectifiers: Surpassing Human-Level Performance on ImageNet Classification}, 2 2015.

\bibitem{Mishkin2015AllInit}
Dmytro Mishkin and Jiri Matas.
\newblock {All you need is a good init}, 11 2015.

\bibitem{Rathi2020EnablingBackpropagation}
Nitin Rathi, Gopalakrishnan Srinivasan, Priyadarshini Panda, and Kaushik Roy.
\newblock {Enabling Deep Spiking Neural Networks with Hybrid Conversion and Spike Timing Dependent Backpropagation}, 5 2020.

\bibitem{Rathi2020DIET-SNN:Networks}
Nitin Rathi and Kaushik Roy.
\newblock {DIET-SNN: Direct Input Encoding With Leakage and Threshold Optimization in Deep Spiking Neural Networks}, 8 2020.

\bibitem{Zimmer2019TechnicalPyTorch}
Romain Zimmer, Thomas Pellegrini, Srisht~Fateh Singh, and Timothée Masquelier.
\newblock {Technical report: supervised training of convolutional spiking neural networks with PyTorch}, 11 2019.

\bibitem{Yin2020EffectiveNetworks}
Bojian Yin, Federico Corradi, and Sander~M. Boht{\'{e}}.
\newblock {Effective and Efficient Computation with Multiple-timescale Spiking Recurrent Neural Networks}, 5 2020.

\bibitem{Fang2021IncorporatingNetworks}
Wei Fang, Zhaofei Yu, Yanqi Chen, Timothée Masquelier, Tiejun Huang, and Yonghong Tian.
\newblock {Incorporating Learnable Membrane Time Constant to Enhance Learning of Spiking Neural Networks}.
\newblock In {\em Proceedings of the IEEE International Conference on Computer Vision}, pages 2641--2651. Institute of Electrical and Electronics Engineers Inc., 2021.

\bibitem{Kim2021OptimizingSensing}
Youngeun Kim and Priyadarshini Panda.
\newblock {Optimizing Deeper Spiking Neural Networks for Dynamic Vision Sensing}.
\newblock {\em Neural Networks}, 144:686--698, 12 2021.

\bibitem{Zheng2021GoingNetworks}
Hanle Zheng, Yujie Wu, Lei Deng, Yifan Hu, and Guoqi Li.
\newblock {Going Deeper With Directly-Trained Larger Spiking Neural Networks}, 2021.

\bibitem{Kim2020RevisitingScratch}
Youngeun Kim and Priyadarshini Panda.
\newblock {Revisiting Batch Normalization for Training Low-latency Deep Spiking Neural Networks from Scratch}, 10 2020.

\bibitem{Ioffe2015BatchShift}
Sergey Ioffe and Christian Szegedy.
\newblock {Batch Normalization: Accelerating Deep Network Training by Reducing Internal Covariate Shift}, 2 2015.

\bibitem{GuoRecDis-SNN:Networks}
Yufei Guo, Xinyi Tong, Yuanpei Chen, Liwen Zhang, Xiaode Liu, Zhe Ma, and Xuhui Huang.
\newblock {RecDis-SNN: Rectifying Membrane Potential Distribution for Directly Training Spiking Neural Networks}.

\bibitem{Guo2023RMP-Loss:Networks}
Yufei Guo, Xiaode Liu, Yuanpei Chen, Liwen Zhang, Weihang Peng, Yuhan Zhang, Xuhui Huang, and Zhe Ma.
\newblock {RMP-Loss: Regularizing Membrane Potential Distribution for Spiking Neural Networks}, 8 2023.

\bibitem{Hu2024SNNResNet}
Jiakui Hu, Tianxiang Hu, Yifan Xu, Zhaokun Zhou, Yonghong Tian, Bo~Xu, and Guoqi Li.
\newblock Advancing spiking neural networks toward deep residual learning.
\newblock {\em arXiv preprint arXiv:2404.03663}, 2024.

\bibitem{Shi2024SpikingResformer}
Xinyu Shi, Zecheng Hao, and Zhaofei Yu.
\newblock Spikingresformer: Bridging resnet and vision transformer in spiking neural networks.
\newblock {\em arXiv preprint arXiv:2403.12345}, 2024.

\bibitem{Zhou2023Spikingformer}
Chenlin Zhou, Liutao Yu, Zhaokun Zhou, Han Zhang, Zhengyu Ma, Huihui Zhou, and Yonghong Tian.
\newblock Spikingformer: Spike-driven residual learning for transformer-based spiking neural network.
\newblock {\em arXiv preprint arXiv:2304.11954}, 2023.

\bibitem{Bellec2018LongNeurons}
Guillaume Bellec, Darjan Salaj, Anand Subramoney, Robert Legenstein, and Wolfgang Maass.
\newblock {Long short-term memory and learning-to-learn in networks of spiking neurons}, 3 2018.

\bibitem{ZenkeTheNetworks}
Friedemann Zenke and Tim~P Vogels.
\newblock {The remarkable robustness of surrogate gradient learning for instilling complex function in spiking neural networks}.

\bibitem{Herranz-Celotti2022StabilizingTraining}
Luca Herranz-Celotti and Jean Rouat.
\newblock {Stabilizing Spiking Neuron Training}, 2 2022.

\bibitem{Ding2022AcceleratingInitialization}
Jianhao Ding, Jiyuan Zhang, Zhaofei Yu, and {Huang Tiejun}.
\newblock {Accelerating Training of Deep Spiking Neural Networks with Parameter Initialization}, 2022.

\bibitem{Perez-Nieves2023SpikingCollapse}
Nicolas Perez-Nieves and Dan F.~M Goodman.
\newblock {Spiking Network Initialisation and Firing Rate Collapse}, 5 2023.

\bibitem{Eshraghian2021TrainingLearning}
Jason~K. Eshraghian, Max Ward, Emre Neftci, Xinxin Wang, Gregor Lenz, Girish Dwivedi, Mohammed Bennamoun, Doo~Seok Jeong, and Wei~D. Lu.
\newblock {Training Spiking Neural Networks Using Lessons From Deep Learning}, 9 2021.

\bibitem{GuoReducingNetworks}
Yufei Guo, Yuanpei Chen, Liwen Zhang, Yinglei Wang, Xiaode Liu, Xinyi Tong, Yuanyuan Ou, Xuhui Huang, and Zhe Ma.
\newblock {Reducing Information Loss for Spiking Neural Networks}.

\bibitem{Fang2021DeepNetworks}
Wei Fang, Zhaofei Yu, Yanqi Chen, Tiejun Huang, Timothée Masquelier, and Yonghong Tian.
\newblock {Deep Residual Learning in Spiking Neural Networks}, 2 2021.

\bibitem{Andrews1998SpecialEngineers}
L.C. Andrews.
\newblock {\em {Special Functions of Mathematics for Engineers}}.
\newblock SPIE Optical Engineering Press, 1998.

\bibitem{Pellegrini2020Low-activityRecognition}
Thomas Pellegrini, Romain Zimmer, and Timothée Masquelier.
\newblock {Low-activity supervised convolutional spiking neural networks applied to speech commands recognition}, 11 2020.

\bibitem{Wu2020DeepRecognition}
Jibin Wu, Emre Yılmaz, Malu Zhang, Haizhou Li, and Kay~Chen Tan.
\newblock {Deep Spiking Neural Networks for Large Vocabulary Automatic Speech Recognition}.
\newblock {\em Frontiers in Neuroscience}, 14, 3 2020.

\bibitem{Liu2021Event-basedNetworks}
Qianhui Liu, Dong Xing, Huajin Tang, De~Ma, and Gang Pan.
\newblock {Event-based Action Recognition Using Motion Information and Spiking Neural Networks}, 2021.

\bibitem{Lecun1998Gradient-basedRecognition}
Y.~Lecun, L.~Bottou, Y.~Bengio, and P.~Haffner.
\newblock {Gradient-based learning applied to document recognition}.
\newblock {\em Proceedings of the IEEE}, 86(11):2278--2324, 1998.

\bibitem{xiao2017fashionmnist}
Han Xiao, Kashif Rasul, and Roland Vollgraf.
\newblock Fashion-mnist: A novel image dataset for benchmarking machine learning algorithms, 2017.

\bibitem{krizhevsky2009learning}
Alex Krizhevsky.
\newblock Learning multiple layers of features from tiny images, 2009.

\bibitem{orchard2015converting}
Garrick Orchard, Ajinkya Jayawant, Gregory~K Cohen, and Nitish~V Thakor.
\newblock Converting static image datasets to spiking neuromorphic datasets using saccades, 2015.

\bibitem{Roy2019TowardsComputing}
Kaushik Roy, Akhilesh Jaiswal, and Priyadarshini Panda.
\newblock {Towards spike-based machine intelligence with neuromorphic computing}.
\newblock {\em Nature}, 575(7784):607--617, 11 2019.

\bibitem{Kingma2014Adam:Optimization}
Diederik~P. Kingma and Jimmy Ba.
\newblock {Adam: A Method for Stochastic Optimization}, 12 2014.

\end{thebibliography}

\end{document}